\documentclass{article}

\PassOptionsToPackage{numbers,sort&compress}{natbib}
\usepackage[final,dblblindworkshop]{neurips_2026}
\workshoptitle{Towards Test-Time Continual Learning Agents}

\usepackage[utf8]{inputenc}
\usepackage[T1]{fontenc}
\usepackage{hyperref}
\usepackage{url}
\usepackage{amsmath,amssymb}
\usepackage{booktabs}
\usepackage{graphicx}
\usepackage{xcolor}
\usepackage{microtype}
\usepackage{caption}
\usepackage{subcaption}
\usepackage{enumitem}
\usepackage{float}

\newcommand{\Learned}{\textsc{Learned}}
\newcommand{\Control}{\textsc{Control}}
\newcommand{\Wrong}{\textsc{Wrong}}
\newcommand{\Fresh}{\textsc{Fresh}}
\newcommand{\LmC}{\ensuremath{\Learned\!-\!\Control}}
\newcommand{\LmW}{\ensuremath{\Learned\!-\!\Wrong}}
\newcommand{\WmC}{\ensuremath{\Wrong\!-\!\Control}}

\title{Binding-Specific Relearning Savings for\\
Test-Time Continual Learning}

\author{%
  Zhe Zhou\\
  Department of Computer Science\\
  University of Washington\\
  \texttt{zhou1106@cs.washington.edu}
}
\date{}

\begin{document}
\raggedbottom
\maketitle

\begin{abstract}
When a fact becomes harder to elicit from a language model, has the model lost
it, or has it merely become less accessible? Parameter-updating test-time
continual learning agents are usually evaluated by current output accuracy; we
ask whether that measurement captures how prior experience shapes later
updates. We test whether learning a specific
factual binding produces a later relearning advantage beyond familiarity with
its entity, answer vocabulary, format, or query slot. In Qwen2.5-1.5B, we
compare three matched, counterbalanced histories: \Learned{} receives the
correct binding, \Control{} receives matched exposure without it, and \Wrong{}
receives a competing relation-valid answer in the same slot. After an identical
rehearsal-free continual fine-tuning stream, \Learned{} reacquires the target
faster than \Control{} at a prespecified checkpoint of partial forgetting.
\Wrong{} shows no positive advantage over \Control{}, while its comparison
with \Learned{} provides slot-matched corroboration, subject to possible
competitor interference. The advantage remains positive after adjustment for
measured pre-relearning behavior. In a secondary descriptive conversion using
an unpaired never-seen pool, prior learning reduces the optimizer updates
required for restoration by about $18\%$. Current behavioral accuracy can
therefore underestimate the lasting effect of previously acquired experience.
Code is available at
\mbox{\url{https://github.com/zhezhou1106/relearning-savings}}.
\end{abstract}

\section{Introduction}
\label{sec:intro}

Continually learning agents modify their weights after deployment, creating a
stability--plasticity problem: new updates must add knowledge without erasing
what came before
\citep{mccloskey1989catastrophic,kirkpatrick2017overcoming,shi2024survey}.
Evaluation usually measures this balance at the output by asking whether the
model still answers earlier questions correctly
\citep{luo2025empirical,chen2024continual}. Such probes are indispensable for
measuring current behavior, but they do not reveal whether an inaccessible fact
still influences subsequent learning.

For a test-time continual learning (TTCL) agent that updates its parameters
from deployment experience, later updates can make acquired information appear
lost at the output. Whether to replay or reteach that experience depends in
part on whether its earlier acquisition still changes the optimization needed
to recover it. We contribute an evaluation protocol for parameter-updating
test-time continual learning: relearning savings complements current behavioral
accuracy by measuring how prior experience changes the cost of a later refresh.

This distinction matters because a model that partially forgot a fact and one
that never learned it can perform alike before retraining yet respond
differently once optimization resumes. We test the corresponding, deliberately
narrow prediction: under partial behavioral forgetting, a previously learned
factual binding can cost less to reacquire than a matched, never-learned
binding. Our diagnostic is \emph{relearning savings}, a classical measure of
memory missed by recall or recognition
\citep{ebbinghaus1885memory,nelson1978detecting,macleod1988forgotten} that also
helped qualify early catastrophic-interference results and remains useful in
modern networks \citep{hetherington1989catastrophic,mayo2023multitask}.
Interpreting savings, however, requires separating the target association from
familiarity with its entity, answer vocabulary, prompt format, and query slot.

We make that counterfactual explicit using $576$ screened synthetic facts and
three matched histories in the same base model. \Learned{} receives the correct
binding; \Control{} receives matched entity, answer, and format exposure without
the binding; and \Wrong{} receives a relation-valid competing answer in the
same slot. A counterbalanced assignment provides complete within-fact pairing.
We then expose every history to the same rehearsal-free continual fine-tuning
stream and launch independent relearning runs from checkpoints along it. A
separate \Fresh{} pool estimates the cost of acquisition without prior
exposure.

Our contribution is therefore not another demonstration that suppressed
behavior can recover, which prior relearning attacks, representation probes,
and oracle comparisons already establish
\citep{deeb2024unlearning,lee2025distillation,xu2025unlearning}. Instead, we
identify how the history of an individual binding changes its reacquisition.
At the prespecified reachable trough, \Learned{} relearns faster than
\Control{}; \Wrong{} shows no positive advantage over \Control{}, and the
slot-matched \LmW{} comparison corroborates the result while providing an upper
bound because \Wrong{} may suffer competitor interference. The advantage
remains positive after adjustment for measured pre-relearning behavior, and
prior learning reduces restoration cost by about $18\%$ relative to naive
acquisition in a secondary descriptive comparison. We additionally show that
equivalence margins between trained
models must be calibrated against ordinary training-seed variation. These
claims concern one model and training regime and do not identify a parameter-
or activation-level storage mechanism.

\section{Related Work}
\label{sec:related}

\textbf{Behavioral forgetting and recoverability.} Several lines of evidence
show that output-level collapse need not imply irreversible loss.
\citet{zheng2025spurious} attribute much of the decline to lost task alignment
rather than lost knowledge; \citet{davari2022probing} and
\citet{ramasesh2021anatomy} show that representational change and behavioral
forgetting need not coincide; \citet{lo2024relearn} observe pruned
concepts reappearing during retraining; \citet{oneill2026continual} track
invented facts through sequential weight writes and find storage surviving
behavioral failure. Unlearning studies similarly distinguish reversible from
more persistent suppression using relearning and representation-level evidence
\citep{deeb2024unlearning,xu2025unlearning,goel2026auditing}.

\textbf{Continual training dynamics.} Continual evaluation reveals a
\emph{stability gap}: performance on prior material drops sharply at a
distribution shift and partially recovers without replay
\citep{delange2023stabilitygap,harun2024overcoming}, a pattern also present in
LM continual pretraining \citep{guo2024efficient,ibrahim2024simple}.

\textbf{Recovery as a probe.} Relearning is now an established robustness test
for unlearning \citep{deeb2024unlearning,hu2025jogging,qian2025layered}; prompting and
out-of-distribution inputs provide complementary recovery attacks
\citep{lynch2024eight,lucki2025adversarial,shumailov2024ununlearning}.
\citet{lee2025distillation} behavior-match a previously capable model to an
oracle never exposed to the target capability and still find faster recovery.
Concurrent work finds that internal accessibility predicts recovery across
models, although it does not reliably identify which individual facts will
recover \citep{song2026measure}.

\textbf{Synthetic factual testbeds.} Studying acquisition and forgetting of
specific facts requires facts the base model does not already know
\citep{kirchenbauer2025fictionalqa,zucchet2025howdo,chang2024howdo}. Work on
pretraining forgetting also motivates entity-sensitive measures beyond
aggregate perplexity \citep{liao2025exploring}, while broader post-training
evaluations argue for measuring multiple dimensions beyond accuracy
\citep{thede2026captrack}.\ \citet{meng2022locating} formalize factual
associations as links between entity--relation queries and their answers.
Appendix~\ref{app:related} expands this discussion.

\section{Methodology}
\label{sec:method}
\label{sec:formulation}

\textbf{Facts, bindings, and slots.} A fact is a triple $(e,r,a)$. Each
relation $r$ has a fixed set $\mathcal{A}_r$ of $|\mathcal{A}_r|=8$
relation-valid answers. The \emph{binding} is the association
$(e,r)\!\to\!a$; the pair $(e,r)$ is the \emph{slot}. A history can occupy the
slot without holding the correct binding, which is what \Wrong{} does. However,
\Wrong{} also installs a competing relation-valid answer; \WmC{} therefore
combines slot familiarity with competitor interference and is not a pure
estimate of slot familiarity.

\begin{table}[t]
\caption{Notation used throughout.}
\label{tab:notation}
\centering
\footnotesize
\begin{tabular}[c]{llll}
\toprule
Symbol & Meaning & Symbol & Meaning \\
\midrule
$(e,r,a)$ & fact and target binding & $h$ & learning history \\
$k$ & continual-stream step & $u$ & relearning step \\
$\mathcal U$ & early relearning window & $L$ & target log-odds \\
$\Delta L$ & between-history behavioral gap & $g_i^h(u)$ & within-history gain \\
$s_i$ & fact-level savings contrast & $S$ & mean log-odds savings \\
\bottomrule
\end{tabular}
\end{table}

\textbf{Primary metric.} Every answer is multi-token, so
$p_\theta(a\mid e,r)$ is the probability of the full answer string under
teacher forcing, without length normalization. We renormalize these
probabilities over $\mathcal{A}_r$ and take the target log-odds, in nats:
\begin{equation}
L(\theta; e,r,a) = \log \frac{\tilde{p}_\theta(a \mid e,r)}
{1 - \tilde{p}_\theta(a \mid e,r)},
\qquad
\tilde{p}_\theta(a\mid e,r) = \frac{p_\theta(a \mid e,r)}
{\sum_{a_j \in \mathcal{A}_r} p_\theta(a_j \mid e,r)} .
\label{eq:logodds}
\end{equation}
Restricting to $\mathcal{A}_r$ removes format and vocabulary drift, so a change
in $L$ reflects a change in which relation-valid answer the model favors. We
corroborate with Jensen--Shannon (JS) distance, exact match, and answer rank.

\textbf{Behavioral accessibility} is what an output-level probe sees. At stream
step $k$, the gap between histories $h,h'$ on a fact is
$\Delta L(k)=L^{h}(k)-L^{h'}(k)$, measured on templates disjoint from all
training templates. Complete behavioral forgetting would correspond to
$\Delta L(k)\!\to\!0$; our confirmatory checkpoint retains a nonzero gap.

\textbf{Relearning savings} is operationalized by a difference in relearning
trajectories. From the checkpoint at stream step $k$ we run an independent
relearning run per history:
$u$ updates of answer-token-only fine-tuning on relearn-train templates,
evaluated on relearn-eval templates. The gain for fact $i$ under history $h$ is
$g_i^{h}(u)=L_i^{h}(u)-L_i^{h}(0)$, which is within-history and so removes each
history's own starting level. Averaging over the early window
$\mathcal{U}=\{1,2,4,8,16\}$ gives $\bar g_i^{h}$. We truncate at $u=16$ to
measure initial reacquisition before widespread saturation and ceiling effects.
The logarithmically spaced steps give equal weight to learning at successive
update scales; their unweighted mean is less dependent on the arbitrary linear
spacing than an area-under-curve summary and less fragile than choosing one
fixed $u$. Figure~\ref{fig:lcgap} shows that the \LmC{} probability gap is
positive across several early steps, peaks near $u=2$, and approaches zero by
$u\ge16$; the summary is therefore front-loaded but not driven by one isolated
budget. The \emph{log-odds savings} of $h$
over reference $h'$ is
\begin{equation}
s_i = \bar g_i^{h} - \bar g_i^{h'},
\qquad
S = \frac{1}{N}\sum_{i=1}^{N} s_i .
\label{eq:savings}
\end{equation}
Because assignment is counterbalanced, $s_i$ is a paired within-fact contrast.
Thus $S$ is a difference-in-differences estimand: it subtracts each history's
own baseline before contrasting the resulting relearning gains.
The tested prediction is $S>0$ at the prespecified stream step despite a
nonzero residual $\Delta L$. $S$ measures additional early log-odds gain, not
the number of updates saved; the conversion to updates is a separate secondary
analysis (\S\ref{sec:cost}).

In continual-learning terminology, $S$ can be viewed as a reflexive form of forward transfer:
past learning of a fact changes the cost of learning that same fact again.
Unlike standard forward and backward transfer, which are usually expressed as
changes in task accuracy, $S$ measures a change in within-fact acquisition
dynamics \citep{lopezpaz2017gem,kemker2018measuring}.

\section{Experiment}
\label{sec:experiment}
\label{sec:setup}

\textbf{Materials.} The panel is $6\text{ relations}\times 8\text{ answers}
\times 12\text{ entities}=576$ synthetic triples, plus a registry relation used
only for control and filler. Screening against the base model rejected $12.6\%$
of candidates for which the model already ranked the designated answer first,
so no panel fact had its designated answer top-ranked at baseline. The panel
splits into a
\textbf{development} set ($384$ facts) used for all gate calibration and
threshold selection and a \textbf{confirmatory} set ($192$) touched only by
final relearning runs; every primary number below is confirmatory. A separate
\Fresh{} pool of $192$ never-seen screened facts is used only at relearning time.
Five disjoint template families separate history-formation training, behavioral
evaluation, relearn training, relearn evaluation, and registry filler.

\textbf{Model and training.} We use the Qwen/Qwen2.5-1.5B base checkpoint,
released under the Apache 2.0 license \citep{qwen2024qwen25}, with full
fine-tuning in bfloat16, three optimization seeds. Relearning uses learning
rate $1\times10^{-5}$ with the loss restricted to answer tokens, at budgets
$u\in\{0,1,2,4,8,16,32,64\}$. Each relearning run carries over checkpoint
weights only: a new AdamW optimizer is initialized for every fact and history,
the learning rate is constant with no warmup, scheduler, or weight decay, and
each update is one full-batch step on the fact's two relearning statements, so
data order cannot differ between histories. The continual stream is rehearsal-free
(${\sim}3$k new facts, no replay of Phase-A targets) and identical across
histories within a seed; we sample it at
$k\in\{0,25,50,75,100,125,150,200,250,400\}$. The confirmatory checkpoint
$k=100$ was selected from this grid by the search reported in
\S\ref{sec:forgetting}.

\textbf{Histories and counterbalancing.} \Learned{} and \Control{} are matched
on entity exposure, answer-token exposure, template family, and total
optimization steps, differing only in whether the binding is ever stated;
\Wrong{} additionally matches the slot, training a relation-valid incorrect
answer obtained by a fixed cyclic shift within $\mathcal{A}_r$. Two-history runs use
a complementary swap (each fact is \Learned{} in one run, \Control{} in the
other); three-history runs use a Latin square (each fact occupies each history
exactly once per seed group). Thus all $192$ confirmatory facts contribute to
every three-history contrast with complete pairing.
Appendix~\ref{app:example} gives the training statements for one fact.

\textbf{Joint gating.} History formation stops jointly: all runs in a seed's
group must pass manipulation checks simultaneously at a shared checkpoint
before it is saved. Gates require \Learned{} exact match $\ge0.90$, \Control{}
registry exact match $\ge0.90$, log-odds separation $\ge1.0$, and \Control{}
residual log-odds $\le-0.5$; the three-history design adds \Wrong{} exact match on
the incorrect answer $\ge0.90$ and \Wrong{} residual log-odds on the true
answer $\le-1.0$. All seeds reached the joint gate in both designs
(Appendix~\ref{app:gates}). Without joint gating, savings could reflect unequal
optimization rather than history.

\section{Result}
\label{sec:results}

\subsection{Forgetting at the reachable trough is partial}
\label{sec:forgetting}

\begin{figure}[t]
\centering
\begin{subfigure}[t]{0.48\textwidth}
\vspace{0pt}
\centering
\includegraphics[width=\textwidth]{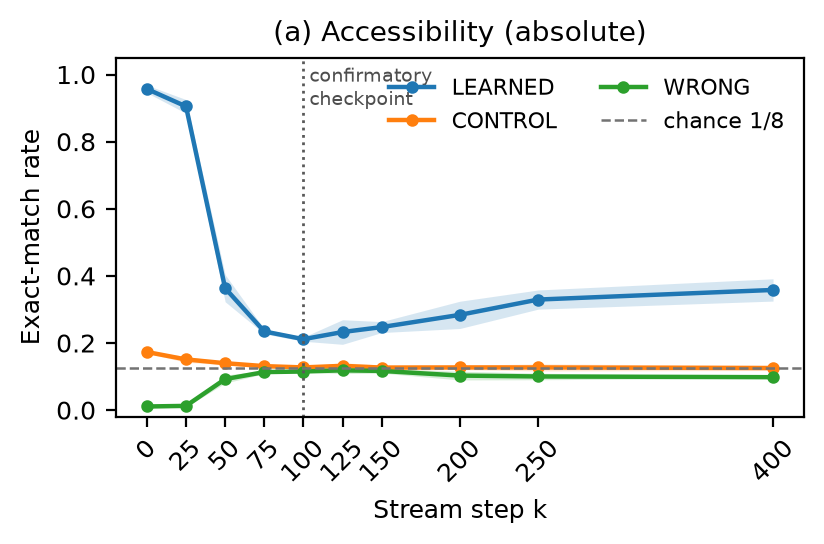}
\phantomsubcaption\label{fig:accessibility}
\end{subfigure}\hfill
\begin{subfigure}[t]{0.48\textwidth}
\vspace{0pt}
\centering
\includegraphics[width=\textwidth]{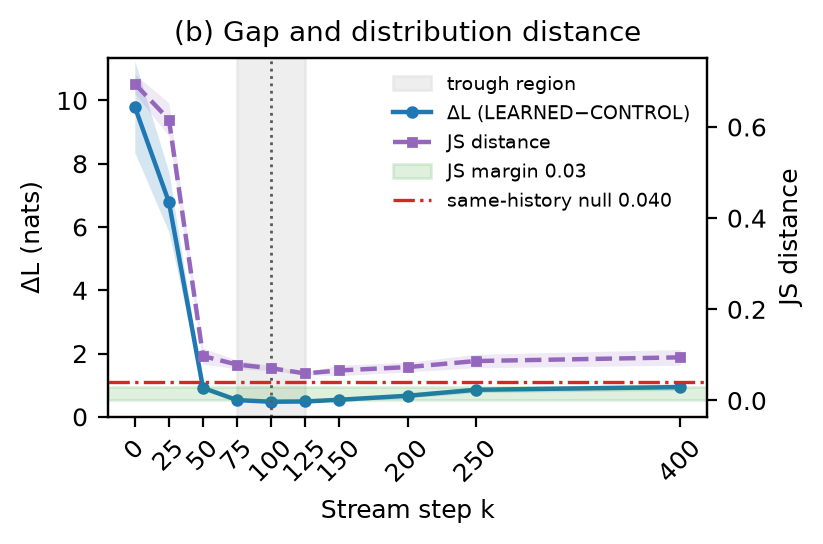}
\phantomsubcaption\label{fig:gapjs}
\end{subfigure}
\caption{Behavioral forgetting along the continual stream. \textbf{(a)} Exact
match for the three histories; the dotted line marks the prespecified
$k=100$ checkpoint and the dashed line marks chance. \textbf{(b)} The
\LmC{} log-odds gap and JS distance. The shaded region contains the observed
trough, the green band is the prespecified JS margin, and the red line is the
same-history cross-seed reference. The histories remain only partially
behavior-matched at the reachable trough.}
\label{fig:forgetting}
\end{figure}

Under the identical rehearsal-free stream, the advantage held by \Learned{}
over \Control{} collapses early and then partially rebounds without replay ---
the stability gap, reproduced at the granularity of individual bindings. Where
the trough sits depends on how densely we look. A coarse checkpoint grid misses
it entirely and places the minimum at $k=250$ with $\Delta L\approx1.1$--$1.5$.
Dense sampling relocates it to $k\approx75$--$125$ with
$\Delta L\approx0.38$--$0.43$ without the destructive perplexity collapse
described below. We call $k=100$ the
\emph{reachable trough}: the prespecified checkpoint at which forgetting is
deepest across seeds among the recipes and checkpoints we searched, subject to
the stream staying healthy (Figure~\ref{fig:forgetting}). All confirmatory
relearning analyses use it, so everything below concerns partial, not complete,
forgetting.

We also searched bounded interventions for a recipe that reaches deeper.
Overwriting with a wrong answer at an elevated learning rate does drive
$\Delta L$ to ${\approx}0.09$, but only by pushing \emph{both} histories to
chance and sending terminal stream perplexity to ${\sim}1900$. The histories
match there because the model has stopped working, so savings measured at that
point would mean nothing. We exclude that recipe.

Two consequences follow, within the recipes and checkpoints we searched.
Behavioral forgetting bottoms out at a nonzero $\Delta L$, so we cannot
condition savings on exact behavioral equivalence. The $\Delta L$ and JS
troughs also fall at different stream steps and never enter their prespecified
margins at the same time, which makes the planned conjunctive equivalence gate
unusable. At the reachable trough \Learned{} still holds a behavioral advantage
of $\Delta L\approx0.38$--$0.43$ nats over \Control{}. We therefore report
savings under partial forgetting and adjust for the residual behavior we can
measure in \S\ref{sec:baseline}. Results at the remaining checkpoints are
descriptive.

The failed gate does not mean the histories never became similar enough. The
prespecified JS margin sits below the distance between models that share a
history and differ only in optimization seed, so \S\ref{sec:jsfloor} treats the
failure as a calibration result rather than as evidence about the histories.

\subsection{Previously learned bindings relearn faster}
\label{sec:savings}

\begin{figure}[t]
\centering
\begin{subfigure}[t]{0.48\textwidth}
\centering
\includegraphics[width=\textwidth]{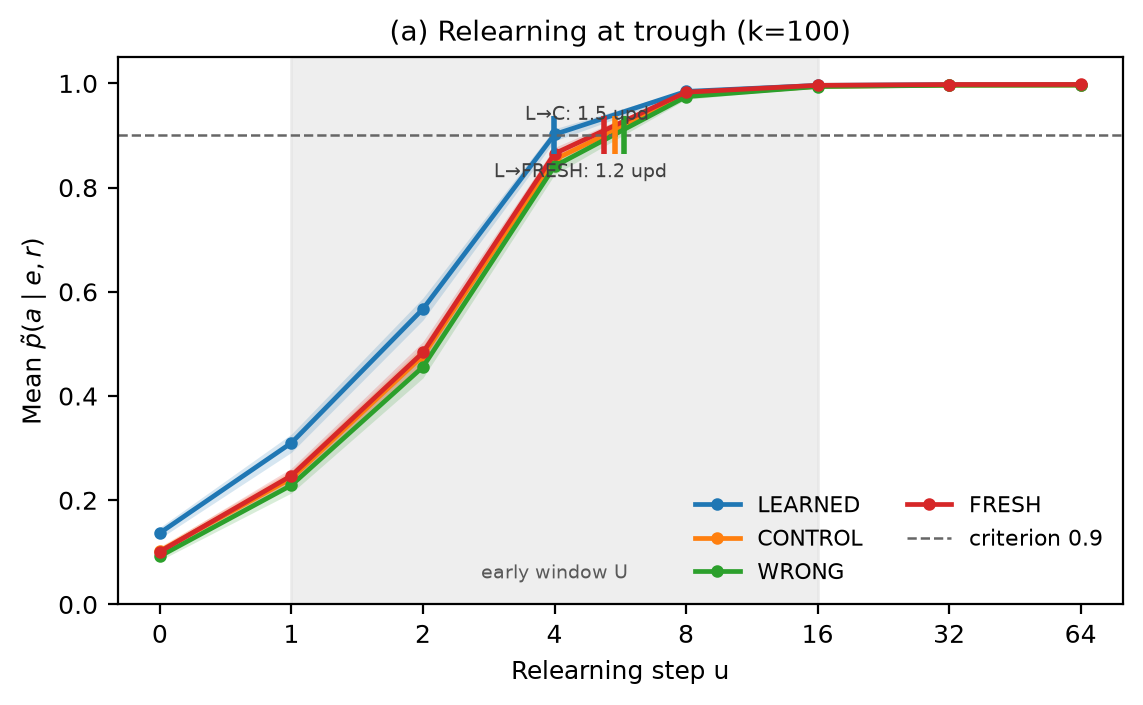}
\phantomsubcaption\label{fig:relearncurves}
\end{subfigure}\hfill
\begin{subfigure}[t]{0.48\textwidth}
\centering
\includegraphics[width=\textwidth]{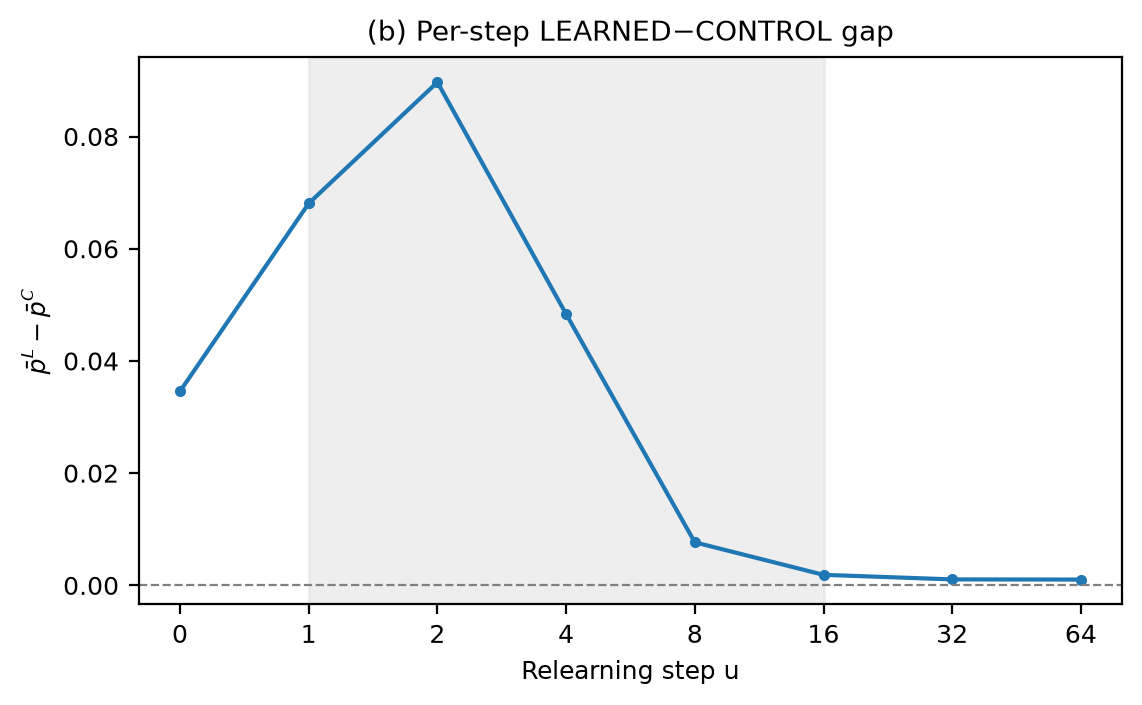}
\phantomsubcaption\label{fig:lcgap}
\end{subfigure}
\caption{Relearning at the reachable trough ($k=100$). \textbf{(a)} Mean
renormalized target probability $\tilde p_\theta(a\mid e,r)$ versus relearning
step $u$ for \Learned{}, \Control{}, and \Wrong{} from the three-history
confirmatory analysis ($N=192$), plus the \Fresh{} pool under the same
relearning protocol. Curves pool seeds 0--2; bands are fact-level bootstrap
$95\%$ intervals. The dashed line is the $0.9$ criterion, the shaded region is
the primary early window $\mathcal U=\{1,2,4,8,16\}$, vertical ticks mark
interpolated crossings, and labels report \Learned-to-\Control{} and
\Learned-to-\Fresh{} update savings. \textbf{(b)} Per-step probability gap
$\bar p^{\Learned}(u)-\bar p^{\Control}(u)$ on the same relearning-step axis.
The advantage is largest early and approaches zero as the histories saturate,
so $S$ summarizes a front-loaded trajectory rather than a uniform lift across
budgets.}
\label{fig:relearning}
\end{figure}

This is our primary result. At the reachable trough ($k=100$),
previously learned bindings reacquire faster than exposure-matched
never-learned bindings in every seed: the \LmC{} contrast gives mean $S=0.177$
nats, with all three bootstrap intervals
excluding zero (Table~\ref{tab:primary-main}). The two histories are matched on
entities, answer vocabulary, template families, optimization budget, and the
stream itself; the only difference is whether the binding was ever stated. The
next two subsections ask whether anything other than the binding could still
account for the gap. Figure~\ref{fig:relearncurves} shows the corresponding
absolute relearning trajectories and criterion crossings.

\begin{table}[t]
\caption{Prespecified \LmC{} relearning-savings result at $k=100$
($N=192$).}
\label{tab:primary}
\label{tab:primary-main}
\centering
\small
\begin{tabular}[c]{rrc}
\toprule
Seed & $S$ (nats) & $95\%$ CI \\
\midrule
0 & $0.160$ & $[0.075,\ 0.249]$ \\
1 & $0.209$ & $[0.061,\ 0.350]$ \\
2 & $0.162$ & $[0.048,\ 0.266]$ \\
\textbf{Mean} & $\mathbf{0.177}$ & \textbf{3/3 pass} \\
\bottomrule
\end{tabular}
\end{table}

Across the descriptive checkpoints from $k=50$ to $k=250$, mean $S$ remains in
the $0.18$--$0.21$ range (Appendix~\ref{app:dose}).

\subsection{Testing slot familiarity as an alternative explanation}
\label{sec:threearm}

The prespecified \LmC{} estimate remains the primary result, but slot
familiarity is an alternative explanation for it. \Learned{} was trained in the
$(e,r)$ slot and \Control{} was not, so \Learned{} may simply be more practiced
at answering that query, whatever the answer. The three-history rebuild adds
\Wrong{}, which occupies the slot with a competing relation-valid answer, and
uses \WmC{} as a negative control: if slot training alone produced a relearning
advantage, \WmC{} should be positive. Because assignment is a Latin square, the
three contrasts satisfy $s_i(\LmW)=s_i(\LmC)-s_i(\WmC)$ exactly for every fact,
so only two are independent.

We detect no positive \WmC{} advantage (mean $-0.002$, 0/3), so the rebuild
gives no evidence that slot training on its own explains the primary \LmC{}
result. This is not a claim that \Wrong{} and \Control{} relearn identically:
in update-savings units the per-seed \WmC{} intervals are $[-0.762,0.030]$,
$[-0.722,0.742]$, and $[-0.406,-0.002]$ (Appendix~\ref{app:deepest}), and two
admit effects in either direction. The rebuild's own \LmC{} estimate reproduces
the two-history mean closely ($0.170$ versus $0.177$), though only 2/3 of its
conditional intervals are positive.

The slot-matched \LmW{} contrast points the same way and is positive in all
three replicates (mean $0.171$); it compares two histories that both trained
the slot and differ only in which answer they were given. It is an upper bound
rather than a third independent finding, because \Wrong{} carries a competing
answer and any handicap from that competitor mechanically inflates \LmW{}. Such
a handicap is visible: adjusted \WmC{} estimates are uniformly negative
($-0.051$, $-0.054$, $-0.026$), as are update savings in
all three replicates. The positive-CI pattern is unchanged at the seed-specific
troughs (Appendix~\ref{app:deepest}).

\begin{table}[t]
\caption{Three-history contrasts at $k=100$. The primary result remains the
prespecified two-history \LmC{} estimate in Table~\ref{tab:primary}; this rebuild
tests its interpretation. Only two rows are independent because
$\LmW=\LmC-\WmC$ exactly. \WmC{} is the negative control, and \LmW{} is
slot-matched corroboration and an upper bound because \Wrong{} is mildly
handicapped. Fractions are descriptive update savings relative to naive
acquisition (\S\ref{sec:cost}).}
\label{tab:threearm}
\centering
\small
\setlength{\tabcolsep}{3pt}
\begin{tabular}[c]{lrrrrcc}
\toprule
& \multicolumn{3}{c}{Per-seed $S$ (nats)} & & & \\
\cmidrule(lr){2-4}
Contrast & seed 0 & seed 1 & seed 2 & Mean $S$ & CI $>0$ & Frac.\ naive \\
\midrule
\LmC{} \emph{(rebuild)} & $0.239$ & $0.189$ & $0.081$ & $0.170$ & $2/3$ & $19\%$ (3/3) \\
\WmC{} \emph{(negative control)} & $0.022$ & $0.009$ & $-0.036$ & $-0.002$ & $0/3$ & $-5\%$ (0/3) \\
\LmW{} \emph{(corroboration)} & $0.217$ & $0.180$ & $0.117$ & $\mathbf{0.171}$ & $\mathbf{3/3}$ & $\mathbf{24\%}$ (3/3) \\
\bottomrule
\end{tabular}
\end{table}

\subsection{The advantage survives adjustment for residual behavior}
\label{sec:baseline}

Because forgetting at the trough is only partial, a second alternative
explanation remains. The within-history subtraction in Eq.~\ref{eq:savings}
removes each history's own starting level, but not the mismatch between the two
starting levels. At $k=100$ \Learned{} still starts higher than \Control{}, and
that residual baseline gap is larger than the raw savings estimate, as the
$u=0$ intercepts in Figure~\ref{fig:relearncurves} make visible. If early gain
simply increased with starting level, savings would be an accounting identity
rather than a finding.

Two sensitivity analyses test this. The first regresses fact-level savings
$s_i$ on the pre-relearn behavioral baseline $d_i$, measured on behavioral-eval
templates disjoint from the relearn-eval templates used to compute $s_i$. Its
intercept at $d=0$ estimates savings for facts with \emph{no} measured residual
behavioral advantage; intercepts are positive in every fit and, in two of three
seeds, larger than the raw estimate (Table~\ref{tab:baseline}). The second
compares per-history LOWESS \citep{cleveland1979lowess} fits of gain against
baseline at matched baselines, and also passes 3/3 while exceeding the raw
estimate.

Slopes are negative or near zero in four of the six fits across the two designs
(the exceptions are two-history seed 2 at $+0.007$ and three-history seed 0 on \LmC{}
at $+0.190$). The intercept claim does not depend on the slope sign and we rest
nothing on it.

Neither analysis is decisive on its own. The intercept can extrapolate when
zero lies outside the common support of the two histories, and the LOWESS
comparison likewise depends on adequate common support. What we claim is
therefore the narrow version: the estimate stays positive after adjustment for
the pre-relearning behavior we can measure. We do not claim that every residual
behavioral difference has been removed.

\begin{table}[t]
\caption{Adjustment for measured baseline behavior at $k=100$ (two-history).
``Residual contrast'' is the LOWESS matched-baseline contrast. The intercept
evaluates the fitted regression at zero residual behavioral advantage and may
involve extrapolation.}
\label{tab:baseline}
\centering
\small
\begin{tabular}[c]{rrrrrc}
\toprule
Seed & Raw $S$ & Intercept at $d=0$ & Slope & Residual contrast & CI $>0$ \\
\midrule
0 & $0.160$ & $0.238$ & $-0.181$ & $0.222$ & yes \\
1 & $0.209$ & $0.228$ & $-0.039$ & $0.264$ & yes \\
2 & $0.162$ & $0.159$ & $+0.007$ & $0.248$ & yes \\
\bottomrule
\end{tabular}
\end{table}

\subsection{How much acquisition cost prior learning removes}
\label{sec:cost}

Nats of log-odds are not what anyone budgets, so as a secondary analysis we
convert savings into optimizer updates: how much of the work of teaching the
fact from scratch does prior learning remove? The denominator comes from the
\Fresh{} pool, which applies the identical relearning protocol to $192$
never-seen screened facts. Its Kaplan--Meier median time to reach
$\tilde p_\theta(a\mid e,r)=0.9$ from a standing start is $3.7$--$4.0$ steps,
interpolated between $u=2$ and $u=4$, the widest gap in the log-spaced budget
grid (linearly in $\log_2(u+1)$; Appendix~\ref{app:repro}).
The $0.9$ threshold was fixed on the development partition as the highest value
that $\ge70\%$ of facts reach in both histories. \Fresh{} facts are disjoint
from the panel by construction, so no per-fact alignment is possible and every
fraction uses the scalar \Fresh{} median at the matching stream step.

At the reachable trough, the primary \LmC{} contrast removes ${\approx}0.69$
relearning steps, or ${\sim}18\%$ of naive acquisition cost, with descriptive
estimates of $24.4\%$ at $k=50$ and $20.7\%$ at $k=150$. The three-history
rebuild agrees: \LmC{} saves $18.6\%$ (3/3 positive fraction intervals), the
slot-matched \LmW{} upper bound saves $23.6\%$ (3/3), and the \WmC{} negative
control saves $-5.0\%$ (0/3). The absolute step counts are small because these
facts are cheap to teach in the first place. Figure~\ref{fig:relearncurves}
gives the visual counterpart in its criterion crossings and \Fresh{}
trajectory. The fraction uses the same fixed $1\times10^{-5}$
answer-token-only relearning protocol for both the panel and \Fresh{} pool and
is not a
recipe-invariant property of the facts or model.

These percentages are less tightly identified than the paired $S$ estimand,
because \Fresh{} is a separate and unpaired pool. In the two-history design,
seed 1's fraction interval at $k=100$ includes zero, although its point
estimate and update-savings direction are both positive. The three-history
fractions borrow the two-history \Fresh{} denominator rather than using a
separate three-history pool; since that denominator is common to all three
contrasts, it rescales them identically and preserves their ordering. Per-seed
tables are in Appendix~\ref{app:deepest}.

\subsection{The planned equivalence gate was uncalibrated}
\label{sec:jsfloor}

The cleanest version of our claim would have conditioned on prespecified
behavioral equivalence, reporting savings at a checkpoint where both $\Delta L$
and JS distance fall inside calibrated margins. We designed for that criterion
and then abandoned it, because JS never entered its prespecified margin of
$0.03$ at any stream step under any non-destructive recipe (minimum
${\approx}0.042$). We first read this as a training failure. It is instead a
margin-calibration failure, and it is quantifiable.

The right reference is the same-history cross-seed JS distance: two models that
share a learning history and differ only in optimization seed sit a minimum
mean distance of $0.040$ apart (CI $[0.039,0.041]$). Our $0.03$ margin lies
below that floor, so it would reject even the same-history comparisons we
measured. A margin that strict cannot support the cross-history equivalence
claim it was meant to license, however the training went.

A third number is easy to confuse with this one. The between-template JS
distance of $0.191$, over $34{,}560$ pairs, measures prompt sensitivity within
a single model, which is a different question. It is not a second estimate of
the cross-seed null and should not be compared against $0.040$ as if the two
were interchangeable; what it does show is that any JS equivalence statement
must specify the template family over which the distance is defined.

\section{Discussion}
\label{sec:discussion}

What these results describe is a dissociation inside a single fact. At the
reachable trough the model's behavior on a learned binding has fallen most of
the way toward its never-learned counterpart, yet the two histories separate
again as soon as optimization resumes. In the terms of
\S\ref{sec:formulation}, prior learning of a fact changes the cost of learning
that same fact again within the early relearning window. The advantage is
front-loaded: Figure~\ref{fig:lcgap} shows it peaking near $u=2$ and approaching
zero by $u\ge16$, while the update-based conversion is less than one optimizer
step. This reflexive forward transfer is invisible to an
output-level probe, which reads the model's current state rather than its
response to further updates. How much of that separation belongs to the
binding itself rather than to what surrounds it is bounded by the controls of
\S\ref{sec:threearm} and \S\ref{sec:baseline}.

The shape of the effect along the stream is informative in its own right.
Savings does not fade as behavioral forgetting deepens. It remains in the
$0.18$--$0.21$ range from $k=50$ through $k=250$, while $\Delta L$ reaches
$0.38$--$0.43$ around $k=75$--$125$ before rebounding, and weakens only late,
once \Learned{} has recovered enough behavioral
advantage for the gain ceiling to bind again. Over that range, how much the
model has forgotten at the output does not tell you how much of the binding's
learning value remains. This is a null over an observed range rather than a
claim of invariance: the range is bounded at both ends by ceiling artifacts of
the gain parameterization, and three seeds cannot exclude a shallow decline.

The contribution is measurement and identification, not the first demonstration
of rapid recovery. Prior work already establishes recoverability after
unlearning, including against a behavior-matched never-exposed oracle
\citep{lee2025distillation,xu2025unlearning}. What our design adds is an answer
to what that recovery advantage is specific to at the level of an individual
factual association. For parameter-updating TTCL agents, the practical
implication is that current output behavior and future learning dynamics answer
different questions: accuracy measures accessibility, while matched-history
relearning measures how prior experience changes subsequent optimization.
Reporting both keeps an accuracy decline from being read as complete loss of a
binding's optimization value. The \Fresh{} median of $3.7$--$4.0$ steps describes a separate never-seen
pool, not the cost of every degraded item; against that reference, the
${\approx}0.69$-step \LmC{} saving suggests that a previously learned binding
may need a shorter refresh than first-time teaching.

The same-history JS analysis carries a second measurement lesson. Equivalence
margins between trained models should be calibrated against ordinary
training-seed variation; ours was not, which is why we make no claim of
behavioral equivalence between histories. The seed floor is a lower bound on a
usable margin; it does not by itself say how large a behavioral difference is
practically negligible. The point generalizes past this
study. Output-distribution distances between independently trained models carry
seed-level variation that an equivalence margin has to account for, so anyone
designing such a criterion should measure the same-history reference
distribution before fixing a margin, or use a difference-in-trajectory estimand
as we do here. The same caution applies to our own negative control: the
absence of a positive \WmC{} result is not evidence that \Wrong{} and
\Control{} are equivalent, only that this protocol found no positive slot-only
advantage.

\section{Limitations}
\label{sec:limitations}

The limits are real and worth stating plainly. All results come from one model
at one scale with full fine-tuning, and the literature reports different
directions for how forgetting changes with model and pretraining scale
\citep{luo2025empirical,ramasesh2022scale}. The facts are synthetic by
necessity, since the design requires bindings the base model does not know, and
synthetic bindings may be encoded differently from naturally acquired ones. The
reachable forgetting frontier of \S\ref{sec:forgetting} is a property of the
recipes we searched, not a proof that no recipe reaches deeper. Fact-level
bootstrap intervals condition on each trained model, and three seeds are too
few to give a population-level training-seed interval. The \Fresh{} conversion
is unpaired, and its three-history denominator is borrowed from the two-history
experiment. The weakest rebuild \LmC{} result, seed 2, is visible in
Table~\ref{tab:threearm}. \Wrong{} confounds slot familiarity with competitor
interference; a control that exposes the slot without consistently teaching a
competing answer is left to future work. Finally, nothing here locates a storage mechanism in
parameters or activations.

\section{Conclusion}
\label{sec:conclusion}

Under partial behavioral forgetting, a factual binding the model once learned
is cheaper to reacquire than a matched binding it never learned. The
competing-answer history shows no net positive advantage over \Control{}, but
because that contrast combines slot familiarity with competitor interference,
it neither establishes equivalence nor rules out a slot contribution. The
overall pattern is consistent with binding-specific savings after matching
entity, answer-vocabulary, and format exposure. The advantage stays positive
after adjustment for measured residual behavior, and in optimizer-update terms
prior learning descriptively removes about $18\%$ of the cost of teaching the
fact from scratch in a secondary, unpaired comparison. In this controlled
setting, current output behavior and future learning
dynamics therefore provide distinct information about the effects of continual
updates, and an evaluation that reports only the first can mistake reduced
accessibility for complete loss.

\paragraph{Broader impacts.}
More accurate retention measurements could reduce unnecessary replay and help
audit continually learning systems. Conversely, relearning savings can reveal
that information remains recoverable after apparent suppression, which matters
for privacy, safety, and machine-unlearning claims. Output-level forgetting
should therefore not be treated as evidence of deletion without recovery tests
and controls matched to the deployment setting. We use synthetic facts and do
not release a new trained model or personal data.

\begin{ack}
We thank the H2Lab at the University of Washington
(\url{https://h2lab.cs.washington.edu/}) for GPU compute.
\end{ack}

\newpage
\bibliographystyle{plainnat}
\bibliography{refs}

\begin{thebibliography}{44}
\providecommand{\natexlab}[1]{#1}
\providecommand{\url}[1]{\texttt{#1}}
\expandafter\ifx\csname urlstyle\endcsname\relax
  \providecommand{\doi}[1]{doi: #1}\else
  \providecommand{\doi}{doi: \begingroup \urlstyle{rm}\Url}\fi

\bibitem[Chang et~al.(2024)Chang, Park, Ye, Yang, Seo, Chang, and
  Seo]{chang2024howdo}
Hoyeon Chang, Jinho Park, Seonghyeon Ye, Sohee Yang, Youngkyung Seo, Du-Seong
  Chang, and Minjoon Seo.
\newblock How do large language models acquire factual knowledge during
  pretraining?
\newblock In \emph{Advances in Neural Information Processing Systems
  (NeurIPS)}, 2024.

\bibitem[Chen et~al.(2026)Chen, Geng, Bhaskar, Friedman, and
  Chen]{chen2024continual}
Howard Chen, Jiayi Geng, Adithya Bhaskar, Dan Friedman, and Danqi Chen.
\newblock Continual memorization of factoids in language models.
\newblock \emph{Transactions on Machine Learning Research}, 2026.

\bibitem[Cleveland(1979)]{cleveland1979lowess}
William~S. Cleveland.
\newblock Robust locally weighted regression and smoothing scatterplots.
\newblock \emph{Journal of the American Statistical Association}, 74\penalty0
  (368):\penalty0 829--836, 1979.
\newblock \doi{10.1080/01621459.1979.10481038}.

\bibitem[Davari et~al.(2022)Davari, Asadi, Mudur, Aljundi, and
  Belilovsky]{davari2022probing}
MohammadReza Davari, Nader Asadi, Sudhir Mudur, Rahaf Aljundi, and Eugene
  Belilovsky.
\newblock Probing representation forgetting in supervised and unsupervised
  continual learning.
\newblock In \emph{IEEE/CVF Conference on Computer Vision and Pattern
  Recognition (CVPR)}, pages 16712--16721, 2022.

\bibitem[De~Lange et~al.(2023)De~Lange, van~de Ven, and
  Tuytelaars]{delange2023stabilitygap}
Matthias De~Lange, Gido~M. van~de Ven, and Tinne Tuytelaars.
\newblock Continual evaluation for lifelong learning: Identifying the stability
  gap.
\newblock In \emph{International Conference on Learning Representations
  (ICLR)}, 2023.

\bibitem[Deeb and Roger(2024)]{deeb2024unlearning}
Aghyad Deeb and Fabien Roger.
\newblock Do unlearning methods remove information from language model weights?
\newblock \emph{arXiv preprint arXiv:2410.08827}, 2024.

\bibitem[Ebbinghaus(1885)]{ebbinghaus1885memory}
Hermann Ebbinghaus.
\newblock \emph{{\"U}ber das Ged{\"a}chtnis: Untersuchungen zur experimentellen
  Psychologie}.
\newblock Duncker \& Humblot, Leipzig, 1885.
\newblock English translation: \emph{Memory: A Contribution to Experimental
  Psychology}, Dover, New York, 1964.

\bibitem[French(1999)]{french1999catastrophic}
Robert~M. French.
\newblock Catastrophic forgetting in connectionist networks.
\newblock \emph{Trends in Cognitive Sciences}, 3\penalty0 (4):\penalty0
  128--135, 1999.
\newblock \doi{10.1016/S1364-6613(99)01294-2}.

\bibitem[Gekhman et~al.(2024)Gekhman, Yona, Aharoni, Eyal, Feder, Reichart, and
  Herzig]{gekhman2024finetuning}
Zorik Gekhman, Gal Yona, Roee Aharoni, Matan Eyal, Amir Feder, Roi Reichart,
  and Jonathan Herzig.
\newblock Does fine-tuning {LLM}s on new knowledge encourage hallucinations?
\newblock In \emph{Proceedings of the 2024 Conference on Empirical Methods in
  Natural Language Processing (EMNLP)}, 2024.

\bibitem[Goel et~al.(2026)Goel, Ritter, and Gurevych]{goel2026auditing}
Anmol Goel, Alan Ritter, and Iryna Gurevych.
\newblock Auditing language model unlearning via information decomposition.
\newblock In \emph{Proceedings of the 19th Conference of the European Chapter
  of the Association for Computational Linguistics (Volume 1: Long Papers)},
  pages 808--826, 2026.
\newblock \doi{10.18653/v1/2026.eacl-long.35}.

\bibitem[Guo et~al.(2024)Guo, Fu, Zhang, Zhao, and Shen]{guo2024efficient}
Yiduo Guo, Jie Fu, Huishuai Zhang, Dongyan Zhao, and Yikang Shen.
\newblock Efficient continual pre-training by mitigating the stability gap.
\newblock \emph{arXiv preprint arXiv:2406.14833}, 2024.

\bibitem[Harun and Kanan(2024)]{harun2024overcoming}
Md~Yousuf Harun and Christopher Kanan.
\newblock Overcoming the stability gap in continual learning.
\newblock \emph{Transactions on Machine Learning Research}, 2024.

\bibitem[Hetherington and Seidenberg(1989)]{hetherington1989catastrophic}
Phil~A. Hetherington and Mark~S. Seidenberg.
\newblock Is there ``catastrophic interference'' in connectionist networks?
\newblock In \emph{Proceedings of the 11th Annual Conference of the Cognitive
  Science Society}, pages 26--33, Hillsdale, NJ, 1989. Lawrence Erlbaum
  Associates.

\bibitem[Hu et~al.(2025)Hu, Fu, Wu, and Smith]{hu2025jogging}
Shengyuan Hu, Yiwei Fu, Zhiwei~Steven Wu, and Virginia Smith.
\newblock Unlearning or obfuscating? jogging the memory of unlearned {LLM}s via
  benign relearning.
\newblock In \emph{International Conference on Learning Representations}, 2025.

\bibitem[Ibrahim et~al.(2024)Ibrahim, Th{\'e}rien, Gupta, Richter, Anthony,
  Lesort, Belilovsky, and Rish]{ibrahim2024simple}
Adam Ibrahim, Benjamin Th{\'e}rien, Kshitij Gupta, Mats~L. Richter, Quentin
  Anthony, Timoth{\'e}e Lesort, Eugene Belilovsky, and Irina Rish.
\newblock Simple and scalable strategies to continually pre-train large
  language models.
\newblock \emph{Transactions on Machine Learning Research}, 2024.

\bibitem[Jones et~al.(2023)Jones, Scott, Ren, Elsayed, Hermann, Mayo, and
  Mozer]{jones2023temporally}
Matt Jones, Tyler~R. Scott, Mengye Ren, Gamaleldin~F. Elsayed, Katherine
  Hermann, David Mayo, and Michael~C. Mozer.
\newblock Learning in temporally structured environments.
\newblock In \emph{International Conference on Learning Representations
  (ICLR)}, 2023.

\bibitem[Kaplan and Meier(1958)]{kaplan1958nonparametric}
Edward~L. Kaplan and Paul Meier.
\newblock Nonparametric estimation from incomplete observations.
\newblock \emph{Journal of the American Statistical Association}, 53\penalty0
  (282):\penalty0 457--481, 1958.
\newblock \doi{10.1080/01621459.1958.10501452}.

\bibitem[Kemker et~al.(2018)Kemker, McClure, Abitino, Hayes, and
  Kanan]{kemker2018measuring}
Ronald Kemker, Marc McClure, Angelina Abitino, Tyler~L. Hayes, and Christopher
  Kanan.
\newblock Measuring catastrophic forgetting in neural networks.
\newblock In \emph{Proceedings of the AAAI Conference on Artificial
  Intelligence}, volume~32, pages 3390--3398, 2018.

\bibitem[Kirchenbauer et~al.(2025)Kirchenbauer, Mongkolsupawan, Wen, Goldstein,
  and Ippolito]{kirchenbauer2025fictionalqa}
John Kirchenbauer, Natjanan Mongkolsupawan, Yuxin Wen, Tom Goldstein, and
  Daphne Ippolito.
\newblock {FictionalQA}: A dataset for studying memorization and knowledge
  acquisition.
\newblock \emph{arXiv preprint arXiv:2506.05639}, 2025.

\bibitem[Kirkpatrick et~al.(2017)Kirkpatrick, Pascanu, Rabinowitz, Veness,
  Desjardins, Rusu, Milan, Quan, Ramalho, Grabska-Barwinska, Hassabis, Clopath,
  Kumaran, and Hadsell]{kirkpatrick2017overcoming}
James Kirkpatrick, Razvan Pascanu, Neil Rabinowitz, Joel Veness, Guillaume
  Desjardins, Andrei~A. Rusu, Kieran Milan, John Quan, Tiago Ramalho, Agnieszka
  Grabska-Barwinska, Demis Hassabis, Claudia Clopath, Dharshan Kumaran, and
  Raia Hadsell.
\newblock Overcoming catastrophic forgetting in neural networks.
\newblock \emph{Proceedings of the National Academy of Sciences}, 114\penalty0
  (13):\penalty0 3521--3526, 2017.
\newblock \doi{10.1073/pnas.1611835114}.

\bibitem[Lee et~al.(2025)Lee, Foote, Infanger, Shor, Kamath, Goldman-Wetzler,
  Woodworth, Cloud, and Turner]{lee2025distillation}
Bruce~W. Lee, Addie Foote, Alex Infanger, Leni Shor, Harish Kamath, Jacob
  Goldman-Wetzler, Bryce Woodworth, Alex Cloud, and Alexander~Matt Turner.
\newblock Distillation robustifies unlearning.
\newblock In \emph{Advances in Neural Information Processing Systems
  (NeurIPS)}, 2025.

\bibitem[Liao et~al.(2025)Liao, Xie, Sun, Sun, and Kang]{liao2025exploring}
Chonghua Liao, Ruobing Xie, Xingwu Sun, Haowen Sun, and Zhanhui Kang.
\newblock Exploring forgetting in large language model pre-training.
\newblock In \emph{Proceedings of the 63rd Annual Meeting of the Association
  for Computational Linguistics (Volume 1: Long Papers)}, 2025.
\newblock \doi{10.18653/v1/2025.acl-long.105}.

\bibitem[Lo et~al.(2024)Lo, Barez, and Cohen]{lo2024relearn}
Michelle Lo, Fazl Barez, and Shay~B. Cohen.
\newblock Large language models relearn removed concepts.
\newblock In \emph{Findings of the Association for Computational Linguistics:
  ACL 2024}, pages 8306--8323, 2024.

\bibitem[Lopez-Paz and Ranzato(2017)]{lopezpaz2017gem}
David Lopez-Paz and Marc'Aurelio Ranzato.
\newblock Gradient episodic memory for continual learning.
\newblock In \emph{Advances in Neural Information Processing Systems},
  volume~30, pages 6467--6476, 2017.

\bibitem[{\L}ucki et~al.(2025){\L}ucki, Wei, Huang, Henderson, Tram{\`e}r, and
  Rando]{lucki2025adversarial}
Jakub {\L}ucki, Boyi Wei, Yangsibo Huang, Peter Henderson, Florian Tram{\`e}r,
  and Javier Rando.
\newblock An adversarial perspective on machine unlearning for {AI} safety.
\newblock \emph{Transactions on Machine Learning Research}, 2025.

\bibitem[Luo et~al.(2023)Luo, Yang, Meng, Li, Zhou, and
  Zhang]{luo2025empirical}
Yun Luo, Zhen Yang, Fandong Meng, Yafu Li, Jie Zhou, and Yue Zhang.
\newblock An empirical study of catastrophic forgetting in large language
  models during continual fine-tuning.
\newblock \emph{arXiv preprint arXiv:2308.08747}, 2023.

\bibitem[Lynch et~al.(2024)Lynch, Guo, Ewart, Casper, and
  Hadfield-Menell]{lynch2024eight}
Aengus Lynch, Phillip Guo, Aidan Ewart, Stephen Casper, and Dylan
  Hadfield-Menell.
\newblock Eight methods to evaluate robust unlearning in {LLM}s.
\newblock \emph{arXiv preprint arXiv:2402.16835}, 2024.

\bibitem[MacLeod(1988)]{macleod1988forgotten}
Colin~M. MacLeod.
\newblock Forgotten but not gone: Savings for pictures and words in long-term
  memory.
\newblock \emph{Journal of Experimental Psychology: Learning, Memory, and
  Cognition}, 14\penalty0 (2):\penalty0 195--212, 1988.
\newblock \doi{10.1037/0278-7393.14.2.195}.

\bibitem[Mayo et~al.(2023)Mayo, Scott, Ren, Elsayed, Hermann, Jones, and
  Mozer]{mayo2023multitask}
David Mayo, Tyler~R. Scott, Mengye Ren, Gamaleldin Elsayed, Katherine Hermann,
  Matt Jones, and Michael~C. Mozer.
\newblock Multitask learning via interleaving: A neural network investigation.
\newblock In \emph{Proceedings of the 45th Annual Conference of the Cognitive
  Science Society (CogSci)}, 2023.

\bibitem[McCloskey and Cohen(1989)]{mccloskey1989catastrophic}
Michael McCloskey and Neal~J. Cohen.
\newblock Catastrophic interference in connectionist networks: The sequential
  learning problem.
\newblock In Gordon~H. Bower, editor, \emph{Psychology of Learning and
  Motivation}, volume~24, pages 109--165. Academic Press, 1989.
\newblock \doi{10.1016/S0079-7421(08)60536-8}.

\bibitem[Meng et~al.(2022)Meng, Bau, Andonian, and Belinkov]{meng2022locating}
Kevin Meng, David Bau, Alex Andonian, and Yonatan Belinkov.
\newblock Locating and editing factual associations in {GPT}.
\newblock In \emph{Advances in Neural Information Processing Systems
  (NeurIPS)}, 2022.

\bibitem[Nelson(1978)]{nelson1978detecting}
Thomas~O. Nelson.
\newblock Detecting small amounts of information in memory: Savings for
  nonrecognized items.
\newblock \emph{Journal of Experimental Psychology: Human Learning and Memory},
  4\penalty0 (5):\penalty0 453--468, 1978.
\newblock \doi{10.1037/0278-7393.4.5.453}.

\bibitem[O'Neill(2026)]{oneill2026continual}
Charles O'Neill.
\newblock Can a language model learn facts continually in its weights?
\newblock \emph{arXiv preprint arXiv:2607.11020}, 2026.

\bibitem[Qian et~al.(2025)Qian, Suriyakumar, Wilson, and
  Hadfield-Menell]{qian2025layered}
Timothy Qian, Vinith Suriyakumar, Ashia Wilson, and Dylan Hadfield-Menell.
\newblock Layered unlearning for adversarial relearning.
\newblock \emph{arXiv preprint arXiv:2505.09500}, 2025.

\bibitem[Ramasesh et~al.(2021)Ramasesh, Dyer, and Raghu]{ramasesh2021anatomy}
Vinay~V. Ramasesh, Ethan Dyer, and Maithra Raghu.
\newblock Anatomy of catastrophic forgetting: Hidden representations and task
  semantics.
\newblock In \emph{International Conference on Learning Representations}, 2021.

\bibitem[Ramasesh et~al.(2022)Ramasesh, Lewkowycz, and Dyer]{ramasesh2022scale}
Vinay~V. Ramasesh, Aitor Lewkowycz, and Ethan Dyer.
\newblock Effect of scale on catastrophic forgetting in neural networks.
\newblock In \emph{International Conference on Learning Representations}, 2022.

\bibitem[Shi et~al.(2024)Shi, Xu, Wang, Qin, Wang, Wang, Wang, Ebrahimi, and
  Wang]{shi2024survey}
Haizhou Shi, Zihao Xu, Hengyi Wang, Weiyi Qin, Wenyuan Wang, Yibin Wang, Zifeng
  Wang, Sayna Ebrahimi, and Hao Wang.
\newblock Continual learning of large language models: A comprehensive survey.
\newblock \emph{ACM Computing Surveys}, 2024.

\bibitem[Shumailov et~al.(2024)Shumailov, Hayes, Triantafillou, Ortiz-Jimenez,
  Papernot, Jagielski, Yona, Howard, and
  Bagdasaryan]{shumailov2024ununlearning}
Ilia Shumailov, Jamie Hayes, Eleni Triantafillou, Guillermo Ortiz-Jimenez,
  Nicolas Papernot, Matthew Jagielski, Itay Yona, Heidy Howard, and Eugene
  Bagdasaryan.
\newblock Ununlearning: Unlearning is not sufficient for content regulation in
  advanced generative {AI}.
\newblock \emph{arXiv preprint arXiv:2407.00106}, 2024.

\bibitem[Song et~al.(2026)Song, Liu, Wang, Li, Li, Gao, Zhang, Lu, Ji, Chang,
  and Chen]{song2026measure}
Zirui Song, Huaxing Liu, Xiang Wang, Shuai Li, Xinye Li, Lang Gao, Jinghui
  Zhang, Zheng Lu, Fengxian Ji, Xiaojun Chang, and Xiuying Chen.
\newblock Measure, don't optimize: Forecasting recovery in {LLM} unlearning.
\newblock \emph{arXiv preprint arXiv:2608.11408}, 2026.
\newblock Concurrent work.

\bibitem[Thede et~al.(2026)Thede, Winzeck, Akata, and
  Schwarz]{thede2026captrack}
Lukas Thede, Stefan Winzeck, Zeynep Akata, and Jonathan~Richard Schwarz.
\newblock {CapTrack}: Multifaceted evaluation of forgetting in {LLM}
  post-training.
\newblock \emph{arXiv preprint arXiv:2603.06610}, 2026.

\bibitem[Xu et~al.(2025)Xu, Yue, Liu, Ye, Hu, and Du]{xu2025unlearning}
Xiaoyu Xu, Xiang Yue, Yang Liu, Qingqing Ye, Haibo Hu, and Minxin Du.
\newblock Unlearning isn't deletion: Investigating reversibility of machine
  unlearning in {LLM}s.
\newblock \emph{arXiv preprint arXiv:2505.16831}, 2025.

\bibitem[Yang et~al.(2024)Yang, Yang, Zhang, Hui, Zheng, Yu, Li, Liu, Huang,
  Wei, et~al.]{qwen2024qwen25}
An~Yang, Baosong Yang, Beichen Zhang, Binyuan Hui, Bo~Zheng, Bowen Yu,
  Chengyuan Li, Dayiheng Liu, Fei Huang, Haoran Wei, et~al.
\newblock Qwen2.5 technical report.
\newblock \emph{arXiv preprint arXiv:2412.15115}, 2024.

\bibitem[Zheng et~al.(2025)Zheng, Cai, Qiu, and Ma]{zheng2025spurious}
Junhao Zheng, Xidi Cai, Shengjie Qiu, and Qianli Ma.
\newblock Spurious forgetting in continual learning of language models.
\newblock In \emph{International Conference on Learning Representations
  (ICLR)}, 2025.

\bibitem[Zucchet et~al.(2025)Zucchet, Bornschein, Chan, Lampinen, Pascanu, and
  De]{zucchet2025howdo}
Nicolas Zucchet, J{\"o}rg Bornschein, Stephanie C.~Y. Chan, Andrew~K. Lampinen,
  Razvan Pascanu, and Soham De.
\newblock How do language models learn facts? dynamics, curricula and
  hallucinations.
\newblock \emph{arXiv preprint arXiv:2503.21676}, 2025.

\end{thebibliography}

\appendix
\newpage
\section{Extended Related Work}
\label{app:related}
Classical work defines savings as faster reacquisition after apparent
forgetting \citep{ebbinghaus1885memory}. Savings can detect memory missed by
recall or recognition, including in comparisons between original and
alternative associations \citep{nelson1978detecting,macleod1988forgotten}. In
neural networks, this measure helped
qualify early catastrophic-interference results
\citep{hetherington1989catastrophic,french1999catastrophic} and has reappeared
in studies of repeated and temporally structured learning
\citep{mayo2023multitask,jones2023temporally}. Reference conditions vary across
this literature, including a learner's earlier trajectory, alternative
associations, and never-exposed oracle models
\citep{nelson1978detecting,macleod1988forgotten,lee2025distillation}.

Recent work on continual learning for language models documents substantial
variation across update regimes, model scales, and evaluation choices
\citep{shi2024survey,luo2025empirical,ibrahim2024simple}. Work on
factual acquisition likewise shows that learning curves depend on how facts are
presented and queried \citep{gekhman2024finetuning,chang2024howdo,
zucchet2025howdo}. Together, these studies show that forgetting and acquisition
estimates are sensitive to the update regime, model scale, presentation format,
and evaluation protocol.

\section{Example Fact and Training Statements}
\label{app:example}
Consider the fact with entity \emph{Myrorak}, relation \emph{element}, and
target answer \emph{zoridium}. Each history trains ten statements in two blocks
of five templates; Table~\ref{tab:example} shows the first template of each
block. \Learned{} states the binding. \Control{} sees the same entity with a
filler relation and the same answer on a different entity, but never the
binding. \Wrong{} fills the same slot with a relation-valid incorrect answer.

\begin{table}[htbp]
\caption{History-formation statements for one fact (first template of each
block).}
\label{tab:example}
\centering
\footnotesize
\begin{tabular}[c]{lp{0.4\textwidth}p{0.4\textwidth}}
\toprule
History & Block 1 & Block 2 \\
\midrule
\Learned{} & In the invented world, Myrorak's element is zoridium. & In the invented world, Myrorak's registry code is R-alpha. \\
\Control{} & In the invented world, Myrorak's registry code is R-alpha. & In the invented world, Dalooq's element is zoridium. \\
\Wrong{} & In the invented world, Myrorak's element is queltrium. & In the invented world, Myrorak's registry code is R-alpha. \\
\bottomrule
\end{tabular}
\end{table}

At relearning, all three histories receive the same two statements,
``Relearn: the element of Myrorak is zoridium.'' and ``Update registry: Myrorak
has element zoridium.'', and are evaluated on the held-out prompts ``Final
check. Myrorak's element?'' and ``Held-out eval. Relation=element;
Entity=Myrorak; Value=''.

\section{History-Formation Gates}
\label{app:gates}
\begin{table}[htbp]
\caption{Joint stopping checkpoints. All runs in a seed's group pass the
manipulation checks simultaneously.}
\centering
\small
\begin{tabular}[c]{rrlrrr}
\toprule
\multicolumn{3}{c}{Two-history} & \multicolumn{3}{c}{Three-history (Latin square)} \\
\cmidrule(r){1-3}\cmidrule(l){4-6}
Seed & Joint stop & Result & Seed & $K^\star$ & Earliest pass (c0/c1/c2) \\
\midrule
0 & $500$ & passed & 0 & $750$ & $750/500/750$ \\
1 & $750$ & passed & 1 & $750$ & $500/750/750$ \\
2 & $750$ & passed & 2 & $500$ & $500/500/500$ \\
\bottomrule
\end{tabular}
\end{table}

\section{Dose--Response Across Stream Checkpoints}
\label{app:dose}
\begin{table}[H]
\caption{Descriptive mean \LmC{} savings across stream step $k$. The $k=100$
comparison is prespecified; all other steps are uncorrected for multiplicity.}
\label{tab:primary-dose}
\centering
\small
\begin{tabular}[c]{rrc}
\toprule
$k$ & Mean $S$ & Seeds passing \\
\midrule
$0$   & $-4.89$ & 0/3 \\
$25$  & $-2.72$ & 0/3 \\
$50$  & $0.205$ & 3/3 \\
$75$  & $0.189$ & 3/3 \\
$\mathbf{100}$ & $\mathbf{0.177}$ & \textbf{3/3} \\
$125$ & $0.202$ & 3/3 \\
$150$ & $0.211$ & 3/3 \\
$200$ & $0.203$ & 3/3 \\
$250$ & $0.177$ & 3/3 \\
$400$ & $0.158$ & 2/3 \\
\bottomrule
\end{tabular}
\end{table}

\section{Seed-Specific and Per-Seed Results}
\label{app:deepest}
\subsection{Seed-specific deepest checkpoints}
Fixing $k=100$ aligns with the prespecified two-history analysis. Using each
seed's deepest $\Delta L$ ($k=100,125,125$) gives \LmC{} values of $0.239$,
$0.145$, and $0.075$ (2/3); \LmW{} values of $0.217$, $0.158$, and $0.105$
(3/3); and \WmC{} values of $0.022$, $-0.012$, and $-0.029$ (0/3).
These descriptive, multiplicity-uncorrected results leave the positive-CI
pattern unchanged; \LmW{} remains an upper bound because \WmC{} is not positive.

\subsection{Per-seed update savings and fractions}
\begin{table}[H]
\caption{Two-history \LmC{}: per-seed relearning-step savings and fraction of naive.}
\centering
\footnotesize
\begin{tabular}[c]{rrrcrcl}
\toprule
Seed & $k$ & Steps saved & $95\%$ CI & Fraction & Fraction CI & CI $>0$ \\
\midrule
0 & $50$  & $0.851$ & $[0.710, 0.997]$ & $0.230$ & $[0.192, 0.270]$ & yes \\
0 & $\mathbf{100}$ & $0.787$ & $[0.358, 1.486]$ & $0.199$ & $[0.090, 0.375]$ & yes \\
0 & $150$ & $0.691$ & $[0.479, 0.957]$ & $0.176$ & $[0.122, 0.243]$ & yes \\
1 & $50$  & $0.984$ & $[0.781, 1.211]$ & $0.270$ & $[0.214, 0.332]$ & yes \\
1 & $\mathbf{100}$ & $0.779$ & $[-0.098, 1.657]$ & $0.209$ & $[-0.026, 0.444]$ & \textbf{no} \\
1 & $150$ & $0.935$ & $[0.532, 1.439]$ & $0.252$ & $[0.143, 0.387]$ & yes \\
2 & $50$  & $0.858$ & $[0.652, 1.065]$ & $0.231$ & $[0.176, 0.287]$ & yes \\
2 & $\mathbf{100}$ & $0.514$ & $[0.304, 0.720]$ & $0.135$ & $[0.080, 0.189]$ & yes \\
2 & $150$ & $0.731$ & $[0.507, 0.953]$ & $0.195$ & $[0.135, 0.254]$ & yes \\
\bottomrule
\end{tabular}
\end{table}

\begin{table}[H]
\caption{Three-history: per-seed relearning-step savings and fraction of naive
at $k=100$. Denominator is the two-history \Fresh{} Kaplan--Meier median
\citep{kaplan1958nonparametric}; \LmW{} is an upper bound.}
\centering
\footnotesize
\begin{tabular}[c]{rlrcrcl}
\toprule
Seed & Contrast & Steps saved & $95\%$ CI & Fraction & Fraction CI & CI $>0$ \\
\midrule
0 & \LmC{} & $0.845$ & $[0.522, 1.175]$ & $0.213$ & $[0.132, 0.297]$ & yes \\
0 & \LmW{} & $1.207$ & $[0.869, 1.567]$ & $0.305$ & $[0.220, 0.396]$ & yes \\
0 & \WmC{} & $-0.362$ & $[-0.762, 0.030]$ & $-0.091$ & $[-0.192, 0.007]$ & no \\
1 & \LmC{} & $0.947$ & $[0.520, 1.744]$ & $0.254$ & $[0.139, 0.467]$ & yes \\
1 & \LmW{} & $0.973$ & $[0.571, 1.571]$ & $0.261$ & $[0.153, 0.421]$ & yes \\
1 & \WmC{} & $-0.027$ & $[-0.722, 0.742]$ & $-0.007$ & $[-0.193, 0.199]$ & no \\
2 & \LmC{} & $0.350$ & $[0.059, 0.623]$ & $0.092$ & $[0.015, 0.163]$ & yes \\
2 & \LmW{} & $0.545$ & $[0.255, 0.809]$ & $0.143$ & $[0.067, 0.212]$ & yes \\
2 & \WmC{} & $-0.195$ & $[-0.406, -0.002]$ & $-0.051$ & $[-0.107, -0.001]$ & no \\
\bottomrule
\end{tabular}
\end{table}

\section{Reproducibility}
\label{app:repro}
The confirmatory partition and the
$\tilde p_\theta(a\mid e,r)=0.9$ threshold were fixed on the development
partition before any confirmatory run. The prespecified two-history \LmC{}
comparison is primary; the three-history rebuild tests whether a positive
slot-only \WmC{} pattern can account for it. The rebuild completed all $90/90$
planned seed--checkpoint--history cells. The continual stream contains
approximately $3{,}000$ new facts and no replay of Phase-A targets.

\textbf{Answer scoring.} All $48$ answers are multi-token under the
Qwen2.5-1.5B tokenizer ($11$ have two tokens, $33$ three, and $4$ four). The
score of an answer is the sum of its token log-probabilities under teacher
forcing given the prompt, with no length normalization; every contrast compares
the same fact, target, and candidate set across histories, so answer length
cancels in the reported estimands. The relearning loss is likewise computed
over all tokens of the answer.

\textbf{Criterion crossing.} Let $p$ be the sigmoid of the target log-odds
averaged over the two relearn-eval templates. Between the two budgets
$u_{\mathrm{lo}}<u_{\mathrm{hi}}$ that bracket the threshold we set
$t=(0.9-p_{\mathrm{lo}})/(p_{\mathrm{hi}}-p_{\mathrm{lo}})$ and
\[
\log_2(u^\star+1)=\log_2(u_{\mathrm{lo}}+1)
+t\,\bigl[\log_2(u_{\mathrm{hi}}+1)-\log_2(u_{\mathrm{lo}}+1)\bigr].
\]
Facts already at or above $0.9$ at $u=0$ are assigned $u^\star=0$, and facts
that never reach $0.9$ are right-censored at $u=64$ in the Kaplan--Meier
estimate.

\textbf{Compute.} All experiments ran on a single NVIDIA L40S GPU, using about
$300$ GPU-hours in total.

\textbf{Code and data.} Code, environment specification, configurations, fact
panels, per-run history assignments, checkpoint manifests, per-seed analysis
outputs, and relearning curves for both designs are available at
\url{https://github.com/zhezhou1106/relearning-savings}, with exact commands to
regenerate the facts, form each history, run the continual stream and
relearning stages, and reproduce every reported analysis, figure, and table
from its source artifact.

\end{document}